\documentclass[runningheads]{llncs}
\usepackage{graphicx}
\usepackage{comment}
\usepackage{amsfonts, amsmath, multirow}
\usepackage[colorlinks = true]{hyperref}
\usepackage[ruled]{algorithm2e}
\newcommand\defeq{\mathrel{\stackrel{\makebox[0pt]{\mbox{\normalfont\tiny def}}}{=}}}

\begin{document}
\title{Practical Aspects of Zero-Shot Learning\thanks{Research was funded in part by the Centre for Priority Research Area Artificial Intelligence and Robotics of Warsaw University of Technology within the Excellence Initiative: Research University (IDUB) programme}}
\author{Elie Saad\inst{1}\orcidID{0000-0002-5627-6351} \and
Marcin Paprzycki\inst{2}\orcidID{0000-0002-8069-2152} \and 
Maria Ganzha\inst{1}\orcidID{0000-0001-7714-4844}}
\authorrunning{E. Saad, et al.}
\institute{Warsaw University of Technology, Warsaw, Poland \\
\email{elie.saad.stud@pw.edu.pl, maria.ganzha@pw.edu.pl}\\
\and
Systems Research Institute Polish Academy of Sciences\\
\email{marcin.paprzycki@ibspan.waw.pl}
}

\maketitle              % typeset the header of the contribution
\begin{abstract}
One of important areas of machine learning research is zero-shot learning. It is applied when properly labeled training data set is not available. A number of zero-shot algorithms have been proposed and experimented with. However, none of them seems to be the ``overall winner''. In situations like this, it may be possible to develop a meta-classifier that would combine ``best aspects'' of individual classifiers and outperform all of them. In this context, the goal of this contribution is twofold. First, multiple state-of-the-art zero-shot learning methods are compared for standard benchmark datasets. Second, multiple meta-classifiers are suggested and experimentally compared (for the same datasets).

\end{abstract}

\section{Introduction}\label{section:introduction}
Classic machine learning methods focus on classifying instances, when the classes have been ``seen'' during training. Whereas, many real world applications require classifying ``entities'' that have not been seen before, e.g., object recognition (where every object is a category), cross-lingual dictionary induction (where every word is a category), etc. Here, one of reasons for this situation is lack of resources to annotate the available (and, possibly, systematically growing) dataset. To solve this problem, zero-shot learning has been proposed. Hence, zero-shot learning can be seen as an example of ``frugal AI'', where the frugality is caused by lack of data annotation, e.g. due to lack of resources.

While large number of zero-shot learning approaches have been proposed (see, for instance,~\cite{lampert2013attribute,larochelle2008zero,rohrbach2011evaluating,yu2010attribute,xu2017matrix,ding2017low,xian2018zero}), they have been experimented with, mostly, in different contexts. Nevertheless, it is possible to conjecture that there is not a single ``best'' approach. In situations like this, meta-classifiers have been proposed. They receive suggestions from individual classifiers and ``judge'', which of them is the ``best''. The assumption is that performance of a meta-classifier will perform better than any of the individual ones.

In this context, the the contribution of this work is twofold. First, a comprehensive comparison of five state-of-the-art zero-shot learning approaches applied to five standard benchmarking datasets. Second, explorations into possibility of developing a meta-classifier for zero-shot learning.

\subsection{Problem Formulation}\label{subsection:problem_formulation}
In zero-shot learning, instances in the feature space are represented as a vector (usually, a real number vector). A subset of training instances is labeled, while testing instances are not. The labeled training classes are referred to as \emph{seen classes}, while the unlabeled testing instances are referred to as \emph{unseen classes}. In this work, the seen and unseen datasets are assumed to be disjoint. However, it is recognized that an interesting area of research covers also the case where some (usually small) overlap between seen and unseen classes exists. 

Let us now formulate the specific problem. Given a dataset of \textit{image embeddings} $\mathcal{X}=\{(x_i,y_i)\in\mathcal{X}\times\mathcal{Y}|i=1,...,N_{tr}+N_{te}\}$, each image is a real $D$-dimensional embedding vector comprised of features $x_i\in\mathbb{R}^D$, and each class label is represented by an integer $y_i\in\mathcal{Y}\equiv\{1,...,N_0,N_0+1,...,N_0+N_1\}$ giving $N_0+N_1$ distinct classes. Here, for generality, it is assumed that $\mathcal{X}\defeq\mathbb{R}^D$. The dataset $\mathcal{X}$ is divided into two subsets: (1) training set and (2) test set. The training set is given by $X^{tr}=\{(x_i^{tr},y_i^{tr})\in\mathcal{X}\times\mathcal{Y}_0|i=1,...,N_{tr}\}$, where $y_i^{tr}\in\mathcal{Y}_0\equiv\{1,...,N_0\}$, resulting in $N_0$ distinct training classes. The test set is given by $X^{te}=\{(x_i^{te},y_i^{te})\in\mathcal{X}\times\mathcal{Y}_1|i=N_{tr}+1,...,N_{te}\}$, where $y_i^{te}\in\mathcal{Y}_1\equiv\{N_0+1,...,N_0+N_1\}$ providing $N_1$ distinct test classes.

The goal of zero-shot learning is to train a model on the training set $X^{tr}$ that performs ``well'' for the test set $X^{te}$. Obviously, zero-shot learning is infeasible without auxiliary information associating labels from the training and the test sets, when the training and test set classes are disjoint $\mathcal{Y}_0\cap\mathcal{Y}_1=\emptyset$. The proposed solution is to represent each class label $y$ $(1\leq y\leq N_0+N_1)$ by its prototype $\pi(y)=p\in\mathcal{P}\defeq\mathbb{R}^M$ (also known as its semantic embedding). Here, $\pi(\cdot):\mathcal{Y}_0\cup\mathcal{Y}_1\rightarrow\mathcal{P}$ is the prototyping function and $\mathcal{P}$ is the semantic embedding space. The prototype vectors are such that any two class labels $y_0$ and $y_1$ are similar if and only if their prototype representations $\pi(y_0)=p_0$ and $\pi(y_1)=p_1$ are close in the semantic embedding space $\mathcal{P}$. For example, their inner product is large in $\mathcal{P}$, i.e, $\langle\pi(y_0),\pi(y_1)\rangle_\mathcal{P}$ is large. Prototyping all of class labels into a joint semantic space, i.e., $\{\pi(y)|y\in\mathcal{Y}_0\cup\mathcal{Y}_1\}$, results in labels becoming related. This takes care of the problem of disjoint class sets, and the model can learn from the labels in the training set, and predict labels from the test set.

\section{Literature Review}\label{section:literature_review}
Many approaches have been proposed for the zero-shot learning. Here, Deep Visual Semantic Embedding (\textit{DeViSE})~\cite{frome2013devise}, Attribute Label Embedding (\textit{ALE})~\cite{akata2015label}, and Structured Joint Embedding (\textit{SJE})~\cite{akata2015evaluation} use a bilinear compatibility function. They follow the \emph{Stochastic Gradient Descent} (SGD), where the SGD~\cite{shalev2014understanding} is implicitly regularized by early stopping. The Embarrassingly Simple Approach to Zero-Shot Learning (\textit{ESZSL})~\cite{romera2015embarrassingly} uses the square loss to learn the bilinear compatibility function and explicitly defines regularization with respect to the Frobenius norm. Kodirov et al. in~\cite{kodirov2017semantic} proposes a semantic encoder-decoder model, called Semantic AutoEncoder (\textit{SAE}), where the training instances are projected into the semantic embedding space $\mathcal{P}$ with the projection matrix $W$, and then projected back into the feature space $\mathcal{X}$, with projection matrix $W^*$ -- where $W^*$ is the conjugate transpose.

Another group of approaches adds a non-linearity component to the linear compatibility function. Here, Latent Embeddings (LATEM)~\cite{xian2016latent} learns a piecewise linear compatibility and uses the same ranking loss function as the \textit{DeViSE}, as well as the SGD. LATEM learns multiple mappings corresponding to different visual characteristics of the data and uses a latent variable to select, which matrix mapping to use in predictions. Cross Modal Transfer (CMT) in~\cite{socher2013zero} trains two-layer neural network using nonlinear mapping function $\tanh$.

A third set of approaches uses probabilistic mapping~\cite{lampert2013attribute}. Here, Direct Attribute Prediction (DAP) algorithm makes class prediction by combining scores of the learned attribute classifiers, whereas the Indirect Attribute Prediction (IAP) algorithm approximates the probabilities of the attributes associated with the training instances, by predicting probabilities of each training class value.

One more group of approaches expresses the input image features and the semantic embeddings as a mixture of seen classes. Here, Semantic Similarity Embedding (SSE)~\cite{zhang2015zero} connects class values within both semantic embedding space and feature embedding space of feature vectors. Convex Combination of Semantic Embeddings (ConSE)~\cite{norouzi2013zero} learns the probability of a feature vector belonging to a specific class. Synthesized Classifiers (SynC)~\cite{changpinyo2016synthesized} first, learns a mapping from the semantic embedding space to the model space, which holds the classifiers associating real class values and phantom class values. The class values along with the set of phantom class values, which are introduced to connect the seen and unseen class values, form a weighted bipartite graph. The objective of SynC is to align the semantic embedding space and the model space, resulting in reducing the distortion error. The Generative Framework for Zero-Shot Learning (GFZSL)~\cite{verma2017simple} applies generative modeling. Here, each class-conditional distribution is modeled as a multivariate Gaussian distribution.

In the fifth set of models, during training, both seen and unseen classes are included in the training data. The Discriminative Semantic Representation Learning (DSRL)~\cite{ye2017zero} can be applied to any  method that is based on a compatibility learning function. It learns the feature embedding vectors from the training instances with non-negative matrix factorization and aligns the feature embedding vectors with the semantic embedding vectors of their corresponding class values.

\section{Selection of Methods and Experimental Setup} \label{section:selection_of_methods_and_experimental_setup}
%
\begin{comment}
In all the experiments presented herein, the classification accuracy of the classifiers is estimated using the Top-1 (T1), Top-5 (T5), Logarithmic Loss (LogLoss), and the F1 score (F1) accuracy measures. Among the many well-known datasets, used in ZSL literature, three \emph{coarse-grained} datasets have been selected: one of which is a small dataset in terms of number of image instances, namely Attribute Apascal\&Ayahoo (aPY)~\cite{farhadi2009describing}, and two of which are large datasets, namely Animals with Attributes 1 (AWA1)~\cite{lampert2013attribute} and Animals with Attributes 2 (AWA2)~\cite{xian2018zero}. Moreover, two \emph{fine-grained} datasets have also been selected, both of which are large: namely Caltech-UCSD-Birds 200-2011 (CUB)~\cite{welinder2010caltech} and Scene UNderstanding (SUN)~\cite{patterson2012sun}. Datasets which contain anywhere between $1$ and $10,000$ ($1\leq N_{tr}+N_{te}\leq10,000$) image instances inclusive are considered as small datasets, whereas any dataset which contains more than $10,000$ ($10,000<N_{tr}+N_{te}$) image instances is considered as a large dataset.
\end{comment}

Based on analysis of the literature five zero-shot learning approaches have been selected: (1) \textit{DeViSE}, (2) \textit{ALE}, (3) \textit{SJE}, (4) \textit{ESZSL}, and (5) \textit{SAE}. This was done, due to the fact that they have been found to be the ``most competitive''.

In the literature, the following datasets have been found to be the most popular benchmarks for zero-shot learning: (a) Caltech-UCSD-Birds 200-2011 \textit{CUB}~\cite{welinder2010caltech}, (b) Animals with Attributes 1 \textit{AWA1}~\cite{lampert2013attribute}, (c) Animals with Attributes 2 \textit{AWA2}~\cite{xian2018zero}, (d) Attribute Apascal\&Ayahoo \textit{aPY}~\cite{farhadi2009describing}, and (e) Scene UNderstanding \textit{SUN}~\cite{patterson2012sun}. Hence, they were selected.

Due to lack of space, state-of-the-art in meta-classifiers has been omitted from the literature review. However, note that this work is focused on zero-shot learning. Hence, five standard meta-classifiers have been tried: (A) Meta-Decision Tree \textit{MDT}~\cite{todorovski2003combining}, (B) deep neural network \textit{DNN}~\cite{goodfellow2016deep}, (C) Game Theory-based approach \textit{GT}~\cite{abreu2006analyzing}, (D) Auction-based model \textit{Auc}~\cite{abreu2006analyzing}, (E) Consensus-based approach \textit{Con}~\cite{alzubi2018consensus}, and (F) a simple majority voting \textit{MV}~\cite{ruta2005classifier}. Here, classifiers (C), (D), (E) and (F) have been implemented directly as in the cited literature. However, the implementation of (A) differs from the one described in~\cite{todorovski2003combining} by not applying a third condition, which is the \emph{weight} condition on the classifiers. This is due to the fact that all individual classifiers are applied simultaneously to the same datasets. Finally, for the \textit{DNN} has been implemented a simple neural network with two hidden layers. The hidden layers and the output layer use the rectified linear activation function. The optimization function is the SGD algorithm, where the mean squared error loss function is used.

The code that has been used in experiments, and the list of hyperparamer values for both the individual classifiers and the meta-classifiers can be found in the Github repository\footnote{\url{https://github.com/Inars/Developing_MC_for_ZSL}}. The hyperparameter values were obtained through a series of ten experiments, after which, the hyperparameter values with the best resulting performance have been selected for the experimentation. We would like to emphasise that the hyperparameter values which were selected are not the best ones to be used, and better accuracy values can be obtained with different hyperparameter values.

The following standard measures have been used to measure the performance of explored approaches: (T1) Top-1, (T5) Top-5, (LogLoss) Logarithmic Loss, and (F1) F1 score. Their definitions can be found in~\cite{xian2018zero,mannor2005cross,sokolova2006beyond}. The reason for the selection of these accuracy measurements is because of their usage abundance in the literature for studying models. 

Finally, the issue of comparing the results of performed experiments with the state-of-the-art reported in~\cite{xian2018zero} has to be addressed. To the best of our knowledge, the codes used there are not publicly available. Thus, the best effort was made to implement the methods using codes based on~\cite{frome2013devise,akata2015label,akata2015evaluation,romera2015embarrassingly,kodirov2017semantic}. The known difference is that the feature vectors and the semantic embedding vectors, provided in the datasets were used, instead of calculated ones. Another difference between the code written for the experimentation and the original codes, is that the dataset splits for the code written follow the proposed splits in~\cite{xian2018zero}, whereas the original codes use the standard splits. Nevertheless, we believe that the results, reported in what follows, fairly represent work reported in~\cite{xian2018zero}.

\section{Experiments with Individual Classifiers}
\label{subsection:experiments_with_individual_classifiers}
The first set of results was obtained using the five individual classifiers applied to the five benchmark datasets. Results displayed in Table~\ref{table:individual_classifier_performance_for_the_top-1_accuracy} show the Top-1 accuracy. They include also those reported in~\cite{frome2013devise,akata2015label,akata2015evaluation,romera2015embarrassingly,kodirov2017semantic} (the \textbf{O} column). The \textbf{R} column represents the replicated results based on~\cite{xian2018zero}. The \textbf{I} columns represents the results of in-house implementations of the five approaches.
\begin{table}[htpb]
\caption[Individual Classifier Performance for the Top-1 Accuracy]{Individual Classifier Performance for the Top-1 Accuracy}
\label{table:individual_classifier_performance_for_the_top-1_accuracy}
\centering
\begin{tabular}{|l|c|c|c|c|c|c|c|c|c|c|c|c|c|c|}
\hline
\multirow{2}{*}{\textbf{CLF}} & \multicolumn{3}{c|}{\textbf{CUB}} & \multicolumn{3}{c|}{\textbf{AWA1}} & \multicolumn{2}{c|}{\textbf{AWA2}} & \multicolumn{3}{c|}{\textbf{aPY}} & \multicolumn{3}{c|}{\textbf{SUN}} \\ \cline{2-15}
& \textbf{O} & \textbf{R}  & \textbf{I} & \textbf{O} & \textbf{R}  & \textbf{I} & \textbf{R}  & \textbf{I} & \textbf{O} & \textbf{R}  & \textbf{I} & \textbf{O} & \textbf{R}  & \textbf{I} \\ \hline
DeViSE & -- & 52 & 46.82 & -- & 54.2 & 53.97 & 59.7 & 57.43 & -- & 37 & 32.55 & -- & 56.5 & 55.42 \\\hline
ALE & 26.3 & 54.9 & \textbf{56.34} & 47.9 & 59.9 & 56.34 & 62.5 & 51.89 & -- & 39.7 & 33.4 & -- & 58.1 & \textbf{62.01} \\\hline
SJE & 50.1 & 53.9 & 49.17 & 66.7 & 65.6 & \textbf{58.63} & 61.9 & \textbf{59.88} & -- & 31.7 & 31.32 & -- & 52.7 & 52.64 \\\hline
ESZSL & -- & 51.9 & 53.91 & 49.3 & 58.2 & 56.19 & 58.6 & 54.5 & 15.1 & 38.3 & \textbf{38.48} & 65.8 & 54.5 & 55.63 \\\hline
SAE & -- & 33.3 & 39.13 & 84.7 & 53.0 & 51.5 & 54.1 & 51.77 & -- & 8.3 & 15.92 & -- & 40.3 & 52.71 \\\hline
\end{tabular}
\end{table}

Looking at column \textbf{O}, it can be seen that the \textit{SAE} individual classifier is most accurate on the \textit{AWA1} dataset ($84.7\%$;\cite{kodirov2017semantic}). Moreover, results for \textit{EZSL} for \textit{SUN} dataset outperform all other classifiers ($65.8\%$;~\cite{romera2015embarrassingly}). 

Since no other results, matching our setup, were available, we will now focus on these reported in column $I$, first. Here, the following observations can be made:
\begin{itemize}
    \item \textit{DeViSE} -- got similar results for \textit{AWA1}, \textit{AWA2}, \textit{aPY}, and \textit{SUN} datasets, whereas there is a $5.18\%$ decrease in accuracy for the \textit{CUB} dataset.
    \item \textit{ALE} -- got similar results for \textit{CUB}, \textit{AWA1}, and \textit{SUN} datasets, whereas there is a $10.61\%$ and a $6.3\%$ decrease in accuracy for \textit{AWA2} and \textit{aPY} datasets.
    \item \textit{SJE} -- got similar results for \textit{CUB}, \textit{AWA2}, \textit{aPY}, and \textit{SUN} datasets, whereas there is a $6.97\%$ decrease in accuracy for the \textit{AWA1}  dataset.
    \item \textit{ESZSL} -- got similar results for all five datasets.
    \item \textit{SAE} -- got similar results for \textit{AWA1} and \textit{AWA2} datasets, whereas there is a $5.83\%$, $7.62\%$ and a $12.41\%$ increase in accuracy for \textit{CUB}, \textit{aPY}, and \textit{SUN} datasets, respectively.
\end{itemize}

Moreover, \textit{ALE} got the highest accuracy value for both the \textit{CUB} and the \textit{SUN} datasets ($56.34\%$ and $62.01\%$), \textit{SJE} got the highest accuracy for both the \textit{AWA1} and the \textit{AWA2} datasets ($58.63\%$ and $59.88\%$), whereas \textit{ESZSL} got the highest accuracy for the \textit{aPY} dataset ($38.48\%$). Note also that neither \textit{DeViSE} nor \textit{SAE} managed to outperform other classifiers for any dataset.  All results are around $50\%$, with the best of all being at $62.01\%$. Finally, it all classifiers performed ``considerably worse'' when applied to the \textit{aPY} dataset. 

Comparing results between columns $R$ and $I$ the first observation is that results are close to each other. Second, depending on the dataset and the method either of the two is slightly more accurate. Overall, out of 25 results, methods based on~\cite{xian2018zero} are more accurate in fifteen cases, while implementations native to this work ``win'' in the remaining ten cases. This being the case, the remaining parts of this contribution are base on our implementation of individual classifiers. However, it should be noted that additional experiments have been performed and the same, general, pattern was observed.

While the Top-1 performance measure has its definite merits -- usually one is interested in performance for the ``key class'' -- other measures also have been used. Therefore, in Table~\ref{table:individual_classifier_performance_for_the_top-5_accuracy} performance of individual classifiers measured using Top-5 accuracy have been presented.
\begin{table}[htbp]
\caption[Individual Classifier Performance for the Top-5 Accuracy]{Individual Classifier Performance for the Top-5 Accuracy}
\label{table:individual_classifier_performance_for_the_top-5_accuracy}
\centering
\begin{tabular}{|l|c|c|c|c|c|}
\hline
\textbf{CLF} & \textbf{CUB} & \textbf{AWA1} & \textbf{AWA2} & \textbf{aPY} & \textbf{SUN} \\ \hline
DeViSE & 51.45 & 65.34 & \textbf{69.03} & \textbf{61.01} & 57.78 \\\hline
ALE & \textbf{64.44} & 63.44 & 63.01 & 60.88 & \textbf{64.24} \\\hline
SJE & 54.27 & \textbf{67.54} & 68.54 & 60.82 & 55.14 \\\hline
ESZSL & 58.25 & 64.91 & 64.52 & 57.93 & 57.92 \\\hline
SAE & 44.97 & 65.56 & 66.98 & 49.29 & 56.18 \\\hline
\end{tabular}
\end{table}

Here, it can be observed that: (i)~\textit{DeViSE} obtained the highest accuracy for the \textit{AWA2} and the \textit{aPY} datasets ($69.03\%$ and $61.01\%$); (ii)~\textit{ALE} reached highest accuracy for \textit{CUB} and \textit{SUN} datasets ($64.44\%$ and $64.24\%$); whereas (iii)~\textit{SJE} obtained highest accuracy for \textit{AWA1} dataset ($67.54\%$). Results of Top-5 accuracy are about $10\%$ higher than the results obtained for the Top-1 accuracy (Table~\ref{table:individual_classifier_performance_for_the_top-1_accuracy}). The best accuracy is close to $70\%$. Again, the \textit{aPY} dataset remains the ``hardest'' also from the perspective of Top-5 measure.

The next set of results, reported in Table~\ref{table:individual_classifier_performance_for_the_logarithmic_loss_accuracy} presents that performance of the individual classifiers assessed form the perspective of Logarithmic Loss Accuracy. 
\begin{table}[htpb]
\caption[Individual Classifier Performance for the Logarithmic Loss Accuracy]{Individual Classifier Performance for the Logarithmic Loss Accuracy}
\label{table:individual_classifier_performance_for_the_logarithmic_loss_accuracy}
\centering
\begin{tabular}{|l|c|c|c|c|c|}
\hline
\textbf{CLF} & \textbf{CUB} & \textbf{AWA1} & \textbf{AWA2} & \textbf{aPY} & \textbf{SUN} \\ \hline
DeViSE & 3.36 & 2.2 & 1.67 & \textbf{2.3} & 3.69 \\\hline
ALE & \textbf{1.87} & 1.88 & 1.92 & 2.53 & 3.68 \\\hline
SJE & 2.93 & \textbf{1.41} & 1.49 & \textbf{2.3} & 3.4 \\\hline
ESZSL & 2.36 & 1.49 & \textbf{1.45} & 7.04 & \textbf{3.36} \\\hline
SAE & 3.6 & 2.06 & 2.03 & 2.51 & 3.92 \\\hline
\end{tabular}
\end{table}

Here, it can be noticed that: (a) \textit{SJE} got the best accuracy for \textit{AWA1} and \textit{aPY} datasets ($1.41$ and $2.3$);  (b) \textit{ESZSL} obtained best accuracy for \textit{AWA2} and \textit{SUN} datasets ($1.45$ and $3.36$), whereas (c) \textit{DeViSE} and \textit{SJE} got the best accuracy for \textit{aPY} dataset ($2.3$). Overall, the \textit{SUN} and \textit{aPY} datasets are the most difficult form the perspective of Logarithmic Loss accuracy.

The final set of results, reported in Table~\ref{table:individual_classifier_performance_for_the_f1_score_accuracy}, presents performance of individual classifiers assessed using F1 Score accuracy. 
\begin{table}[htpb]
\caption[Individual Classifier Performance; F1 Score Accuracy]{Individual Classifier Performance; F1 Score Accuracy}
\label{table:individual_classifier_performance_for_the_f1_score_accuracy}
\centering
\begin{tabular}{|l|c|c|c|c|c|}
\hline
\textbf{CLF} & \textbf{CUB} & \textbf{AWA1} & \textbf{AWA2} & \textbf{aPY} & \textbf{SUN} \\ \hline
DeViSE & 0.47 & 0.48 & \textbf{0.55} & 0.27 & 0.55 \\\hline
ALE & 0.53 & 0.53 & 0.42 & 0.3 & \textbf{0.62} \\\hline
SJE & 0.49 & 0.55 & \textbf{0.55} & \textbf{0.43} & 0.53 \\\hline
ESZSL & \textbf{0.54} & \textbf{0.57} & 0.53 & 0.25 & 0.56 \\\hline
SAE & 0.39 & 0.44 & 0.45 & 0.08 & 0.53 \\\hline
\end{tabular}
\end{table}

In this case it can be seen that: (a) \textit{SJE} is most accurate for \textit{AWA2} and \textit{aPY} datasets ($0.55$ and $0.43$); (b) \textit{ESZSL} is the most accurate for \textit{CUB} and \textit{AWA1} datasets ($0.54$ and $0.57$); (c) \textit{ALE} is the most accurate for \textit{SUN} dataset ($0.62$); (d) \textit{DeViSE} and \textit{SJE} are the most accurate for \textit{AWA2} ($0.55$). Thus, \textit{aPY} dataset, again, is the hardest to master.

Overall, it is clear that (A) different performance measures ``promote'' different zero-shot learning approaches; (B) \textit{aPY} is most difficult to learn from; (C) as expected, none of individual classifiers can be seen as a clear winner. Therefore, a simplistic method has been proposed to gain better understanding of ``overall strength'' of individual classifiers. However, what follows ``should be treated with a grain of salt''. There are many ways to ``combine performance''. 

In the proposed method, individual classifiers obtain points $5$, $4$, $3$, $2$, and $1$ for best, second best, third best, fourth best, and fifth best accuracy obtained for each dataset. This process is repeated for all four accuracy measures. After points were added, they have been reported in Table~\ref{table:individual_classifier_combined_performance}.
\begin{table}[htpb]
\caption[Individual Classifier Combined Performance]{Individual Classifier Combined Performance}
\label{table:individual_classifier_combined_performance}
\centering
\begin{tabular}{|l|c|c|c|c|c|c|}
\hline
\textbf{CLF} & \textbf{CUB} & \textbf{AWA1} & \textbf{AWA2} & \textbf{aPY} & \textbf{SUN} & \textbf{Total} \\ \hline
DeViSE & 8 & 8 & 17 & 16 & 11 & 60 \\\hline
ALE & \textbf{19} & 11 & 7 & \textbf{15} & \textbf{18} & 70 \\\hline
SJE & 12 & \textbf{19} & \textbf{18} & \textbf{15} & 8 & 72 \\\hline
ESZSL & 17 & 14 & 14 & 11 & 17 & \textbf{73} \\\hline
SAE & 4 & 8 & 8 & 7 & 6 & 33 \\\hline
\end{tabular}
\end{table}

Here, it can be observed that: (a) \textit{ALE} performed the best overall on \textit{CUB} and \textit{SUN} da\-ta\-sets; (b) \textit{SJE} performed best overall on \textit{AWA1} and \textit{AWA2} da\-ta\-sets; (c) \textit{SJE} and \textit{ALE} performed best on \textit{aPY} data\-set. The surprising result is that, while \textit{ESZSL} is not a ``winner'' for a single dataset, it is the best overall performer, when combining results from all datasets (73 points). Finally, \textit{SAE} is clearly the weakest method both within individual datasets and for combined performance. Combined performance of \textit{ALE}, \textit{SJE}, and \textit{EZSL} is very similar.

Overall, reported experiments further confirm that not a single individual classifier outperforms others. Experimental results depend on the data and on used accuracy measure. These, in turn, are application dependent. 

\subsection{Analysis of the Datasets}\label{subsection:analysis_of_the_datasetss}
Since the performance of the classifiers is directly related to the datasets, in this section, this relation is further elaborated. Specifically, for each dataset, its difficulty has been explored. Hence, for each instance, it is deemed to be: (a)~\textit{lvl 0} if it has been correctly identified by \textit{no} individual classifier; (b)~\textit{lvl 1} if it has been identified by \textit{one} classifiers; (c)~\textit{lvl 2} if it has been identified by \textit{two} classifiers; (d)~\textit{lvl 3} if it has been identified by \textit{three} classifiers; (e)~\textit{lvl 4} if it has been identified by \textit{four} classifiers; (and f) \textit{lvl 5} if it has been identified by \textit{all} classifiers. The results in Table~\ref{table:analysis_of_instance_difficulty}, show (in percents) how many instances belong to each category, for each dataset.
\begin{table}[htpb]
\caption[Analysis of Instance Difficulty]{Analysis of Instance Difficulty}
\label{table:analysis_of_instance_difficulty}
\centering
\begin{tabular}{|l|c|c|c|c|c|c|}
\hline
\textbf{CLF} & \textbf{lvl 0} & \textbf{lvl 1} & \textbf{lvl 2} & \textbf{lvl 3} & \textbf{lvl 4} & \textbf{lvl 5} \\ \hline
\textbf{CUB} & 23.86 & 15.98 & 13.11 & 12.4 & 15.23 & 19.41 \\ \hline
\textbf{AWA1} & 20.47 & 15.67 & 11.29 & 12.19 & \textbf{19.33} & 21.04 \\ \hline
\textbf{AWA2} & 21.88 & 12.95 & 13.46 & \textbf{14.62} & 12.26 & 22.84 \\ \hline
\textbf{aPY} & \textbf{36.37} & \textbf{25.76} & \textbf{15.75} & 11.75 & 9.53 & 0.85 \\ \hline
\textbf{SUN} & 19.31 & 12.78 & 10.69 & 12.5 & 16.88 & \textbf{27.85} \\ \hline
\end{tabular}
\end{table}

Looking at Table~\ref{table:analysis_of_instance_difficulty}, a number of observations can be made. (1) Difficulty of the \textit{aPY} dataset is clear, as it has largest percents of instances that have not been recognized by a single classifier (36.37\%), or by one or two classifiers only (jointly, 41.51\%). At the same time, only 0.85\% of its instances have been recognized by all classifiers. Overall only about 22\% of its instances can be considered relatively easy to be recognized. These observations further augment analysis of the \textit{aPY} dataset found in~\cite{xian2018zero}. (2) On the other side of the spectrum is the \textit{SUN} dataset, which not only has 27.85\% of instances that have been correctly recognized by all classifiers. It also has about 55\% of instances that can be considered as relatively easy. (3) Other than \textit{aPY}, the remaining datasets have largest numbers of instances that are either most difficult (\textit{lvl 0}) or easiest (\textit{lvl 5}). 

Approaching the issue from yet another side, ``influence'' of all attributes has been ``scored''. For each correctly recognized instance, all its attributes have been given +1 ``point''. For each incorrectly recognized instance, all its attributes were given -1 ``point''. Out of all attributes of correctly (incorrectly) recognized instances (treated separately), these with highest score have been marked as \textit{easiest} (\textit{hardest}). Results of this operation have been reported in Table~\ref{table:analysis_of_the_datasets}. There, \textit{total} refers to the result, in which all individual classifiers have been considered jointly; whereas \textit{all} refers to all individual classifiers (and \textit{total}), treating the same attribute \textit{easiest} (\textit{hardest}).
\begin{table}[htpb]
\caption[Analysis of the Datasets]{Analysis of the Datasets}
\label{table:analysis_of_the_datasets}
\centering
\begin{tabular}{|l|c|c|c|}
\hline
\textbf{CLF} & \textbf{Easiest Attribute} & \textbf{Individual Classifier} & \textbf{Hardest Attribute} \\ \hline
\textbf{CUB} & has eye color black & all & has eye color black \\ \hline
\multirow{4}{*}{\textbf{AWA1}} & old world & DeViSE; SAE & group \\ \cline{2-4}
& fast & ALE & ocean \\ \cline{2-4}
& old world & SJE; ESZSL & ocean \\ \cline{2-4}
& quadrupedal & total & swims \\ \hline
\multirow{2}{*}{\textbf{AWA2}} & old world & DeViSE; ALE; SJE; SAE; total & group \\ \cline{2-4}
& old world & ESZSL & ocean \\ \hline
\multirow{2}{*}{\textbf{aPY}} & metal & DeViSE; ALE; SJE; ESZSL & metal \\ \cline{2-4}
& head & SAE & metal \\ \cline{2-4}
& furry & total & metal \\ \hline
\textbf{SUN} & no horizon & all & man-made \\ \hline
\end{tabular}
\end{table}

Starting off with the \textit{CUB} dataset: it can be seen that the attribute ``has eye color black'' is both the easiest and the hardest attribute. Inspecting images in the dataset, most (if not all) birds in the dataset seem to have that attribute in some form. This means that the ``has eye color black'' attribute will be associated with most instances that have been both correctly and incorrectly classified, regardless of which attribute is responsible for the result reported by individual classifiers. In this way, the ``has eye color black'' attribute can be seen as ``useless'' as it does not ``help'' in distinguishing between classes. Obviously, there may be also other attributes that are equally useless.

Looking at the \textit{AWA1} dataset: it can be see that the attribute ``oldworld'' is the easiest to classify, whereas the attribute that was easiest to classify by all individual classifiers combined ``quadrapedal''. Similarly, the hardest attribute to classify is ``ocean'', whereas the hardest to classify is ``swims''.

Now, moving to the \textit{AWA2} dataset. It is quite similar to the \textit{AWA1} dataset. Here, again, the attribute ``oldworld'' is the easiest to classify. The hardest attribute to classify is ``group''.

In the case of the \textit{aPY} dataset, the easiest and hardest attribute to classify for individual classifiers is ``metal''. The easiest attribute to be classified by DeViSE, ALE, SJE, and ESZSE is ``metal'', by SAE is ``head'', whereas for the five individual classifiers combined is ``furry''. Judging by the images and their correlating prototypes, the attribute ``metal'' seems to be associated with the largest number of instances (similar to the attribute ``has eye color black'' from the \textit{CUB} dataset), thus we come to the same conclusion that ``metal'' is classified incorrectly regardless of whether it was the reason for the misclassification or not.

Finally, the \textit{SUN} dataset, in which the easiest attribute to classify is ``no horizon'', whereas the hardest attribute to classify ``man-made''. Again, looking at the images from the dataset, it can be seen why that is since some instances under some classes such as ``zen garden'' are relatively hard to classify since their prototypes are similar -- considering that they were engineered -- making the attribute ``man-made'' the reason for the misclassification of a lot of instances.

In summary, results reported above lead to an interesting research question. What would happen if the most obvious/popular attributes were excluded from the dataset. However, exploring this question is out of scope of this contribution.

\section{Meta-Classifiers}\label{section:meta-classifiers}

The last part of research concerns usability of basic meta-classifiers, as mechanisms to combine predictions made by individual classifiers.  Before proceedings, let us note that the number of \textit{hard} instances, found for each dataset and reported in Section~\ref{section:selection_of_methods_and_experimental_setup}, establishes the hard ceiling for: \textit{DNN}, \textit{MDT}, and \textit{MV}. Specifically, here, if not a single individual classifier gave a correct answer, their combination will also not be able to do so. Therefore, for these meta-classifiers, for the 5 datasets, the performance limits are \textit{CUB}: 76.13\%, \textit{AWA1}: 79.53\%, \textit{AWA2}: 78.13\%, \textit{aPY}: 63.63\%, and \textit{SUN}: 80.7\%.

In Table~\ref{table:table:meta-classifier_performance;_top-1_accuracy}, performance of the six meta-classifiers is compared using the Top-1 measure, where the \textbf{Best} row denotes the best result obtained from the ``winning'' individual classifier, for given dataset (see, Table~\ref{table:individual_classifier_performance_for_the_top-1_accuracy}).
\begin{table}[htpb]
\caption[Meta-classifier performance; Top-1 accuracy]{Meta-classifier performance; Top-1 accuracy}
\label{table:table:meta-classifier_performance;_top-1_accuracy}
\centering
\begin{tabular}{|l|c|c|c|c|c|}
\hline
\textbf{CLF} & \textbf{CUB} & \textbf{AWA1} & \textbf{AWA2} & \textbf{aPY} & \textbf{SUN} \\ \hline
\textbf{MV} & \textbf{53.43} & \textbf{58.71} & 56.56 & 32.72 & 61.94 \\\hline
\textbf{MDT} & 47.89 & 56.43 & 51.89 & 33.40 & \textbf{62.08} \\\hline
\textbf{DNN} & 48.63 & 57.56 & 54.72 & \textbf{34.89} & 60.63 \\\hline
\textbf{GT} & 46.58 & 56.75 & \textbf{59} & 32.63 & 59.51 \\\hline
\textbf{Con} & 46.82 & 53.97 & 57.43 & 32.55 & 55.42 \\\hline
\textbf{Auc} & 47.89 & 56.34 & 51.89 & 33.40 & 62.01 \\\hline
\textbf{Best} & 56.34 & 58.63 & 59.88 & 38.48 & 62.01 \\\hline
\end{tabular}
\end{table}

Here, \textit{MV} achieved highest accuracy for \textit{CUB} and \textit{AWA1} datasets ($53.43\%$ and $58.71\%$); \textit{MDT} achieved highest accuracy for \textit{SUN} dataset ($62.08\%$); \textit{DNN} got the highest accuracy for \textit{aPY} dataset ($34.89\%$); finally, \textit{GT} reached highest accuracy for \textit{AWA2} dataset ($59\%$). Moreover, neither \textit{Con} nor \textit{Auc} managed to get the best accuracy for a single dataset.

Comparing results obtained by meta-classifiers with the best individual classifiers, one can see that: (a) the best individual classifier performed better than the best meta-classifier on \textit{CUB}, \textit{AWA2}, and \textit{aPY} (2.91\%, 0.88\%, and 3.59\% better); (b) best meta-classifier performed better than the best individual classifier on \textit{AWA1} and \textit{SUN} datasets (0.08\% and 0.08\% better).

The second view of the results, for the meta-classifiers, was obtained when the F1 measure was applied. The results are summarized in Table~\ref{table:meta-classifier_performance;_f1_score_accuracy}.
\begin{table}[htpb]
\caption[Meta-classifier Performance; F1 Score Accuracy]{Meta-classifier Performance; F1 Score Accuracy}
\label{table:meta-classifier_performance;_f1_score_accuracy}
\centering
\begin{tabular}{|l|c|c|c|c|c|}
\hline
\textbf{CLF} & \textbf{CUB} & \textbf{AWA1} & \textbf{AWA2} & \textbf{aPY} & \textbf{SUN} \\ \hline
\textbf{MV} & \textbf{0.54} & \textbf{0.55} & \textbf{0.52} & \textbf{0.34} & \textbf{0.62} \\\hline
\textbf{MDT} & 0.48 & 0.53 & 0.42 & 0.31 & \textbf{0.62} \\\hline
\textbf{DNN} & 0.49 & 0.53 & 0.46 & 0.28 & 0.6 \\\hline
\textbf{GT} & 0.36 & 0.26 & 0.37 & 0.24 & 0.34 \\\hline
\textbf{Con} & 0.47 & 0.26 & 0.36 & 0.27 & 0.55 \\\hline
\textbf{Auc} & 0.48 & 0.53 & 0.42 & 0.31 & \textbf{0.62} \\\hline
\textbf{Best} & 0.54 & 0.57 & 0.55 & 0.43 & 0.62 \\\hline
\end{tabular}
\end{table}

Here, \textit{MV} achieved highest accuracy for \textit{CUB}, \textit{AWA1}, \textit{AWA2}, and \textit{aPY} datasets ($0.54,0.55,0.52$, and $0.34$); \textit{MV}, \textit{MDT}, and \textit{Auc} achieved highest accuracy for \textit{SUN} dataset ($0.62$). Moreover, neither \textit{DNN} nor \textit{GT} nor \textit{Con} managed to get the best accuracy for a single dataset.

Comparing results obtained by meta-classifiers with the best individual classifiers, one can see that: (a) the best individual classifier performed better than the best meta-classifier on \textit{AWA1}, \textit{AWA2}, and \textit{aPY} ($0.02,0.03,$ and $0.09$ better); (b) best meta-classifier performed equally to the best individual classifier, on \textit{CUB} and \textit{SUN} datasets ($0.54$ and $0.62$).

Using the same method of point distribution, used for measuring the ``combined performance'' of the individual classifiers, results reported in Table~\ref{table:meta-classifier_combined_performance} have been obtained (combining both performance measures -- Top-1 and F1). In this case, the top scorer obtained 6 points, as there are six meta-classifiers.
\begin{table}[htpb]
\caption[Meta-Classifier combined performance]{Meta-Classifier combined performance}
\label{table:meta-classifier_combined_performance}
\centering
\begin{tabular}{|l|c|c|c|c|c|c|}
\hline
\textbf{CLF} & \textbf{CUB} & \textbf{AWA1} & \textbf{AWA2} & \textbf{aPY} & \textbf{SUN} & \textbf{Total} \\ \hline
MV & \textbf{12} & \textbf{12} & \textbf{10} & \textbf{10} & 10 & \textbf{54} \\\hline
MDT & 8 & 8 & 6 & \textbf{10} & \textbf{12} & 44 \\\hline
DNN & 10 & 10 & 8 & \textbf{10} & 8 & 46 \\\hline
GT & 4 & 8 & 9 & 5 & 5 & 31 \\\hline
Con & 6 & 5 & 7 & 5 & 5 & 28 \\\hline
Auc & 8 & 7 & 6 & \textbf{10} & \textbf{11} & 42 \\\hline
\end{tabular}
\end{table}

Here it can be noticed that: (a) \textit{MV} performed the best overall (54 points); (b) \textit{Con} performed the worst overall (28 points); (c) \textit{MV}, \textit{MDT}, and \textit{DNN} out performed \textit{GT}, \textit{Con}, and \textit{Auc}, while their performance was quite similar ($42-46$ points).

It should be noted that the ``simplest'' approaches from both the individual classifiers (ESZSL) and the meta-classifiers (MV) obtained the highest number of points using the ``combined performance'' method in  both Table~\ref{table:meta-classifier_combined_performance} and Table~\ref{table:individual_classifier_combined_performance}.

Finally, the ``combination method'' was applied jointly to both the meta-classifiers and the individual classifiers. Here, Top-1 and the F1 accuracy measures were taken into account. Moreover, since 11 classifiers were compared, the top score was 11 points. Table~\ref{table:meta-classifier_and_individual_classifier_combined_performance} displays the resulting total scores.
\begin{table}[htpb]
\caption[Meta-Classifier and individual classifier combined performance]{Meta-Classifier and individual classifier combined performance}
\label{table:meta-classifier_and_individual_classifier_combined_performance}
\centering
\begin{tabular}{|l|c|c|c|c|c|c|}
\hline
\textbf{CLF} & \textbf{CUB} & \textbf{AWA1} & \textbf{AWA2} & \textbf{aPY} & \textbf{SUN} & \textbf{Total} \\ \hline
DeViSE & 13 & 12 & 20 & 12 & 13 & 70 \\\hline
ALE    & \textbf{21} & 15 & 11 & 17 & 21 & 85 \\\hline
SJE    & 17 & \textbf{20} & \textbf{22} & 16 & 10 & 85 \\\hline
ESZSL  & 21 & 16 & 16 & 16 & 15 & 84 \\\hline
SAE    & 10 & 10 & 11 & 7  & 11 & 49 \\\hline
MV     & 20 & 21 & 17 & \textbf{18} & 20 & \textbf{96} \\\hline
MDT    & 15 & 16 & 11 & \textbf{18} & \textbf{22} & 82 \\\hline
DNN    & 17 & 18 & 15 & 17 & 18 & 85 \\\hline
GT     & 10 & 14 & 15 & 11 & 13 & 63 \\\hline
Con    & 13 & 10 & 13 & 12 & 13 & 61 \\\hline
Auc    & 15 & 15 & 11 & \textbf{18} & 21 & 80 \\\hline
\end{tabular}
\end{table}

Here, it can be noticed that: (a)~\textit{MV} performed the best overall (96 points, better by 11 points than the next three classifiers: \textit{ALE}, \textit{SJE}, and \textit{DNN}); (b)~\textit{SAE} performed the worst overall (28 points); (c)~among competitive approaches, one can list also \textit{ESZL} and \textit{MDT}, and \textit{Auc}; (d)~meta-classifiers performed better overall than the individual classifiers (averaging $77.83$ vs. $74.6$ points). Obviously, since meta-classifiers combined results of individual classifiers, there is a lot of ``methodological grey area''. Recognizing this we can still state that when results from multiple individual classifiers are available, then they can be combined using \textit{simple majority} voting to improve overall performance of classification process. At the same time, use of other (more advanced) meta-classifiers is not straightforwardly leading to performance gains.

\section{Concluding Remarks}

In this contribution, performance of five zero-shot learning models has been studied (\textit{DeViSE}, \textit{ALE}, \textit{SJE}, \textit{ESZSL}, and \textit{SAE}), when applied to standard benchmarking datasets. Moreover, the practical nature of the difficulties of these datasets has been explored. Finally, use of six standard meta-classifiers (MV, MDT, DNN, GT, Con, and Auc) has been experimented with.

The main findings were: (1)~there is not a single best classifiers, and their performance depends on the, application-specific, dataset and applied performance measure; (2)~the \textit{aPY} dataset is the most difficult (of the five that have been tried) to achieve good results of zero-shot learning; (3)~use of standard meta-classifiers may brings some overall benefits; (4)~despite explicit performance ceiling, the \textit{MV}, \textit{MDT}, and \textit{DNN} classifiers, they performed better than their counterparts; (5)~simplest methods obtained best results (both the individual classifier \textit{ESZSL} and the meta-classifier \textit{MV}).

Relatively low quality of obtainable results (accuracy of prediction of order of less than 70\%) suggests that a lot of research is needed for both individual zero-shot learning approaches and, possibly, meta-classifiers. Moreover, work should be verified on datasets similar to \textit{aPY}, which are particularly difficult to apply to. Here, some attention should be devoted to the role that individual attributes play in class (instance) recognition. 
\bibliographystyle{splncs04}
\bibliography{references}
\end{document}